\documentclass[letterpaper, 10 pt, conference]{ieeeconf}  

\IEEEoverridecommandlockouts                              
\usepackage{graphics} 
\usepackage{epsfig} 
\usepackage{mathptmx} 
\usepackage{times} 
\usepackage{amsmath} 
\usepackage{amssymb}  
\usepackage{subfigure}
\usepackage{balance}
\usepackage{siunitx}
\usepackage{comment}
\usepackage{xcolor}
\usepackage{dirtytalk}
\usepackage{bm}
\usepackage{gensymb}

\newcommand{\ndn}[1]{\textcolor{blue}{#1 - NDN}}

\title{\LARGE \bf
A Dexterous Tip-extending Robot with Variable-length Shape-locking
}

\author{Sicheng Wang$^{1*}$, Ruotong Zhang$^{2}$, David A. Haggerty$^{1}$, Nicholas D. Naclerio$^{1}$, and Elliot W. Hawkes$^{1}$
\thanks{This work was supported by the NSF grant No. 1637446 and Undergraduate Research and Creative Activities grant of UC Santa Barbara, No.3510. The work of N. Naclerio was supported by a NASA Space Technology Research Fellowship. }
\thanks{$^{1}$Department of Mechanical Engineering, University of California, Santa Barbara, CA 93106}
\thanks{$^{2}$School of Electrical Engineering, Xi'an Jiaotong University, Xi'an, China, 710049}
\thanks{*Email: \tt\small sicheng\_wang@ucsb.edu }
    }

\begin{document}

\maketitle
\thispagestyle{empty}
\pagestyle{empty}

\begin{abstract}

Soft, tip-extending \say{vine} robots offer a unique mode of inspection and manipulation in highly constrained environments. 
For practicality, it is desirable that the distal end of the robot can be manipulated freely, while the body remains stationary.
However, in previous vine robots, either the shape of the body was fixed after growth with no ability to manipulate the distal end, or the whole body moved together with the tip.
Here, we present a concept for shape-locking that enables a vine robot to move only its distal tip, while the body is locked in place.
This is achieved using two inextensible, pressurized, tip-extending, chambers that \say{grow} along the sides of the robot body, preserving curvature in the section where they have been deployed.
The length of the locked and free sections can be varied by controlling the extension and retraction of these chambers.
We present models describing this shape-locking mechanism and workspace of the robot in both free and constrained environments.
We experimentally validate these models, showing an increased dexterous workspace compared to previous vine robots.
Our shape-locking concept allows improved performance for vine robots, advancing the field of soft robotics for inspection and manipulation in highly constrained environments.

\end{abstract}

\section{INTRODUCTION}
Over recent decades, continuum manipulators have emerged as a promising field in robotics~\cite{idwalker}. 
Various designs have been developed, many converging to a standard architecture featuring a flexible, continuously deforming backbone, and several circumferentially arranged actuators that span its length~\cite{drus}.
The flexible backbone is commonly constructed from soft or semi-soft structures, such as springs, polymer rods, or inflatable chambers.
Other backbones use rigid, hyper-redundant links in series.
Actuators are typically tension cables or pneumatic artificial muscles. (See \cite{webster2010design} for a review).

While the compliant body of a continuum robot can potentially have infinite degrees-of-freedom (DOF), its limited number of actuators makes this nominal advantage less relevant.
A continuum manipulator segment that can achieve a controllable constant curvature can only reach a set of points located on a curved surface in 3-dimensional space, each with only a single orientation~\cite{bjones,yang}. 
This workspace is often unfavorable for practical applications.
As a result, most continuum manipulators are composed of multiple constant-curvature segments mounted in series~\cite{meng,idwalker2,jzyang}. 
This configuration, however, can result in a large number of independently controlled actuators that dramatically increase the cost and complexity of the system~\cite{jzyang,gilbert}.
We thus consider an alternative strategy in which a high number of possible active DOFs is achieved with the combination of a simple activating mechanism and minimum number of actuators.

\begin{figure}[t]
    \centering
    \includegraphics[width=.7\columnwidth]{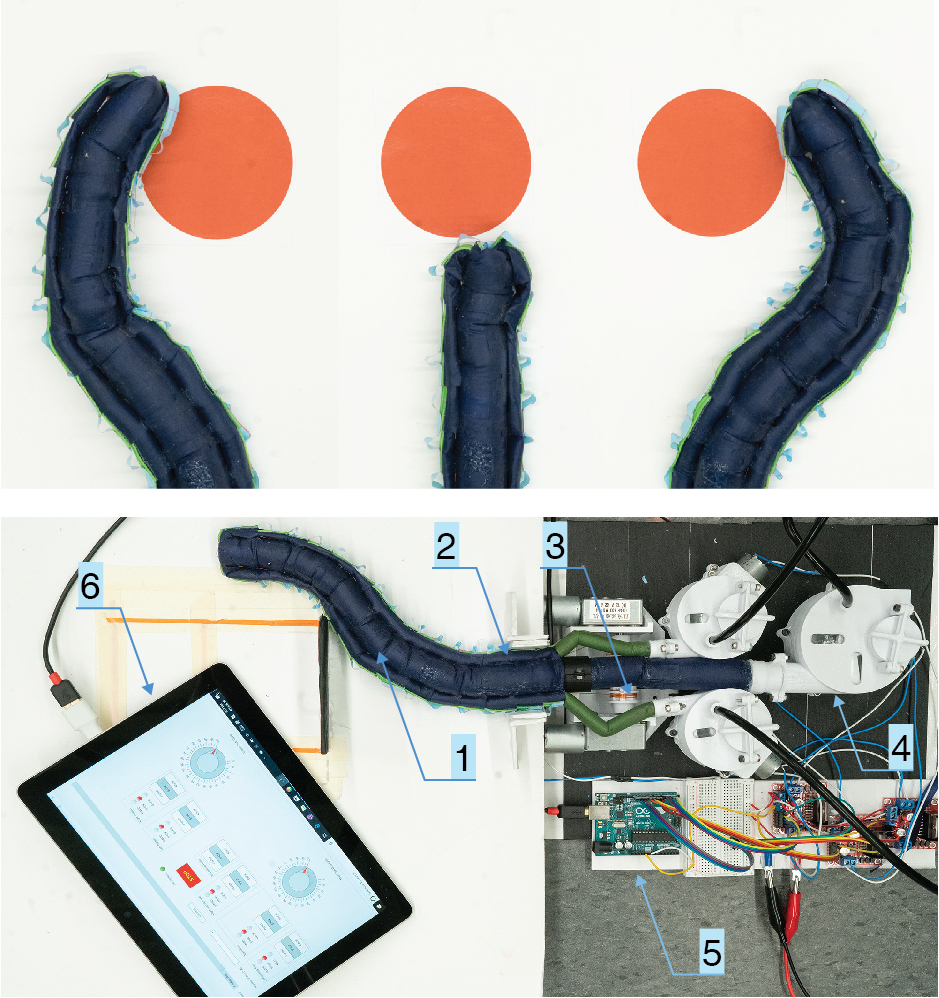}
    \caption{Top: The shape-locking vine robot can achieve compound curvatures with full-length, single curvature actuators by steering, locking, and growing a new unlocked section. This allows it to approach a target from multiple orientations. The sequence of locking and growth could be repeated many times to create complex shapes, and the process is reversible and reconfigurable. Bottom: Composition of the robot. 1: main tip-extending body, 2: shape-locking body (in guiding tube), 3: tendon and reel assembly, 4: reel chamber, 5:microcontroller 6: control terminal on a tablet PC.}
    \label{fig:firstFig}
\end{figure}

The possible active DOFs of a continuum robot can be activated or deactivated by tuning the stiffness of the robot body. While stiffening is typically used to increase strength, here we utilize the mechanism to enable or disable the deflection of a robot segment due to the actuating force and thus activating the possible DOFs. This concept has previously been applied to a modular continuum manipulator\cite{swang2016}.
Multiple mechanisms have been developed for tunable-stiffness manipulators, namely granular jamming~\cite{ranzani,ngcheng}, layer and torsion jamming~\cite{yjkim,sad}, viscosity-tuning of infill medium~\cite{ajloeve,fa}, and antagonistic interactions between backbone and actuators~\cite{ashelge}. 
Some of these designs have demonstrated applicability in scenarios where highly articulated manipulation is desired, most notably in minimally-invasive surgeries~\cite{yjkim,sad,ajloeve,fa}.

Navigation through tip-extension offers an opportunity to improve the dexterity of continuum robots~\cite{ewhawkes}.
Through the eversion of a thin-film material, the distal end of these \say{vine} robots extends, offering a new axial DOF.
This capability has been combined with various mechanisms to aid in navigation.
Such mechanisms include a) non-active,  pre-programming of shape~\cite{lhblu}; b) non-active use of environmental constraints~\cite{jdgreer,luong}; c) active, irreversible, preferential lengthening of the sides of the everting thin-film body~\cite{ewhawkes}; and, d) active steering by cable or artificial muscle~\cite{jdgreer2}.
While (a-c) can achieve complex configurations, they are generally permanent and therefore less applicable for manipulators.
Mechanism (d) is reconfigurable, giving the robot a manipulation capability comparable to classical continuum manipulators, but with an extended workspace due to axial extension and relatively simple control principles.
This however has certain drawbacks, for example a limited range of end effector orientation, and undesirable contact between its proximal sections and the environment during distal manipulation.

In this work, we present a tendon-driven, tip-extending robot that allows a variable-length section of the robot to switch between a stiffened and unstiffened state.
Stiffening takes place from the base up to an arbitrary point along its length, and thus decouples the distal, free-moving segment from the rest of the robot---i.e., locking the shape of the stiffened, proximal segment.
This is achieved by extending two or three smaller-diameter, tip-extending, shape-locking bodies along the robot, which retain the curvature of the robot.
By retracting the shape-locking bodies and the robot, this process is reversible and re-configurable.
Our prototype demonstrates that shape-locking enables a tip-extending manipulator to achieve complex trajectories without the aid of its environment (Fig. \ref{fig:firstFig}).
This produces a higher fidelity recreation of the principle of growth in natural vines than previous tip-extending robots.
It mimics the difference in stiffness between the newly-grown and matured part of a plant vine, which enable them to actively search for resources by deforming the distal end while maintaining a rigid proximal foundation~\cite{gartner}. An alternative strategy for shape-locking, which uses vacuum layer-jamming magnetically activated by a carriage travelling along the inflated robot, has been presented in \cite{bhdo2020}.

This paper is structured as follows: Section \ref{sec:design}
explains the principle of shape locking, Section \ref{sec:fabrication} describes the fabrication of a prototype, Section \ref{sec:model} models the performance and workspace of the robot, Section \ref{sec:results} validates these models, and Section \ref{sec:discussion} discusses the implications of this work.

\section{DESIGN}
\label{sec:design}
The tip-extending, shape-locking robot draws inspiration from certain plant vines, the stems of which stiffen and thus develop a self-supporting structure when an external structure is unavailable.
Poison oak (\textit{Toxicodendron diversilobum}), for example, is known to grow as a shrub, rather a vine, when no external support is present, with its mature stem attaining a higher stiffness than the supple vine.~\cite{gartner}.

\begin{figure}[tb]
  \centering
  \includegraphics[width=0.8\columnwidth]{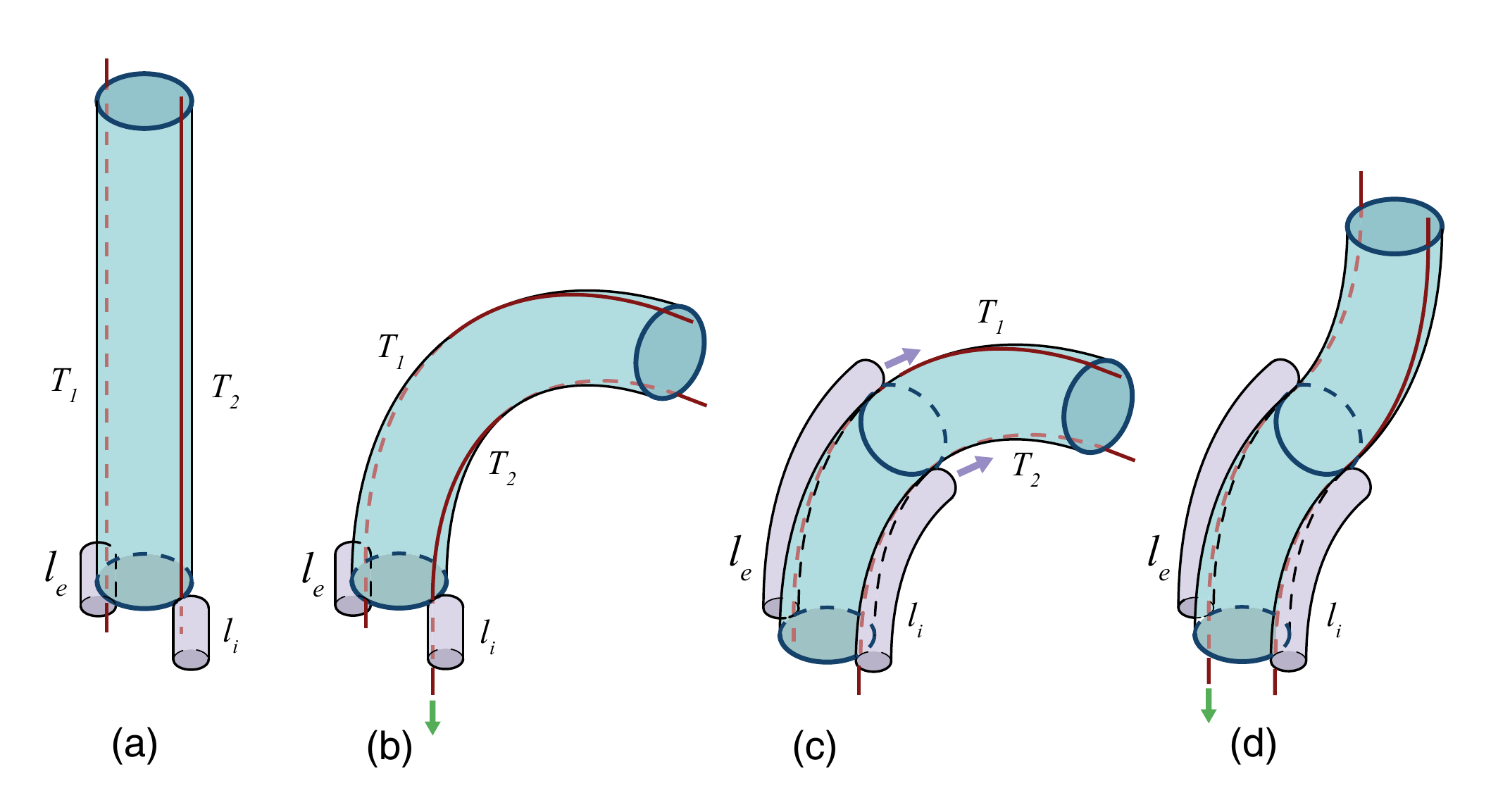}
  \caption{The operation of the shape-locking mechanism. (a): The initial state; (b): tendon (red) $T_2$ is retracted while $T_1$ remains unchanged, resulting in curvature; (c) The locking bodies (purple) grow through the guiding tubes (not shown) to lock in the shape of the curve; (d) by releasing $T_2$ and retracting $T_1$, a second curvature is achieved. Further everting the main body (blue) extends the robot, allowing the above steps to be performed on the newly everted portion.}
  \label{fig:lockingMechanism}
\end{figure}

To achieve a similar effect in a vine robot, we first examine its actuation principles. 
The posture of a planar continuum manipulator is determined by two arcs (the external arc $l_e$, and the internal arc $l_i$) 
located on the two sides of the robot and concentric with the its neutral axis (Fig. \ref{fig:lockingMechanism}). 
Our robot is actuated by varying the relative lengths of $l_e$ and $l_i$ by shortening two tendons embedded 180{\degree} apart along the pneumatic backbone of the robot, one each on $l_e$ and $l_i$.
As a tendon is shortened, that side of the robot contracts, inducing a curvature.
As the tendon is lengthened, the internal pressure of the robot returns it to its neutral position.
Previous works realized this with pneumatic artificial muscles~\cite{jdgreer3}. 

The posture of the robot can be maintained by resisting the change in length of $l_e$ and $l_i$.
We achieve this by the following mechanism.
First, two flexible guiding tubes are attached along the main robot body, over the pull tendons.
Through these tubes evert two shape-locking bodies that extend with the same mechanism as the main body.
Both the inside of the guiding tube and exterior of the shape-locking body are coated with a high friction material, preventing them from sliding relative to each other.
The guiding tubes are split into short, incremental units, such that as the pressurized locking body grows through them, the spacing between them is held constant.
As a result, the curvature of the robot body is held constant, even as tension in the pull tendons change.
See Fig. \ref{fig:lockingMechanism} for an illustration of this principle.

\section{Fabrication}
\label{sec:fabrication}
\subsection{Thin-Film Tip-extending Bodies}
The main body tube, guiding tubes, and shape locking bodies are all constructed as follows.
A long rectangular piece of 0.07 mm thick, silicon-coated ripstop nylon fabric is rolled into a tube and adhered together with a lap joint with a silicone adhesive (Sil-poxy, Smooth-On).
The main body tube is \SI{23}{\milli\meter} in diameter and is sealed together at one end, and attached to a reel chamber at the other, which stores the inverted portion of the robot body.
Nylon tendons are routed through a series of \SI{10}{\milli\meter} long, \SI{0.7}{\milli\meter} diameter catheter segments, spaced \SI{10}{\milli\meter} apart along the length of the main body.
Each tendon is actuated by a set of reel and DC motor installed at the base of the robot.
The 1:1 space-to-length ratio of the catheter segments limits the minimum value of $l_e$ and $l_i$ to $l/2$, preventing extremely small-radius bends through which it is difficult to grow the shape-locking body.
The guiding tube for the shape-locking body is composed of a series of \SI{40}{\milli\meter} long, \SI{11}{\milli\meter} diameter tube segments.
A segment of \SI{1}{\milli\meter} thick polyurethane is attached to the outside of each guiding tube segment to prevent folding that may hinder the extension of the shape-locking body.
Guiding tube segments are attached to the main body with the silicone adhesive.
A long, thin strip of polyurethane film is attached along the length of all the guiding tube segments, to help maintain their relative positions with respect to their neighbors.
The \SI{11}{\milli\meter} diameter shape-locking bodies, and inside of the guiding tubes are coated with cured silicone adhesive to increase friction between the two.

Both the main body and the locking bodies are stowed in rigid, 3D-printed pressurized reels.
Strings attached to their tips are attached to a capstan inside the reel.
A DC motor attached to each aids their retraction.
\begin{figure}[tb]
  \centering
  \subfigure[]{\includegraphics[width = 0.25\columnwidth]{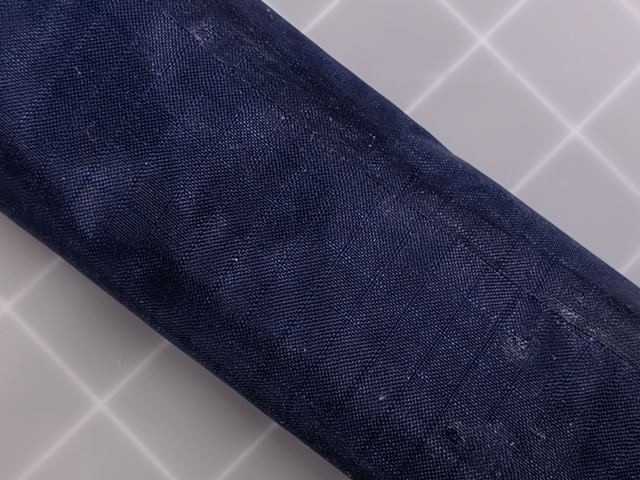}}
  \subfigure[]{\includegraphics[width = 0.25\columnwidth]{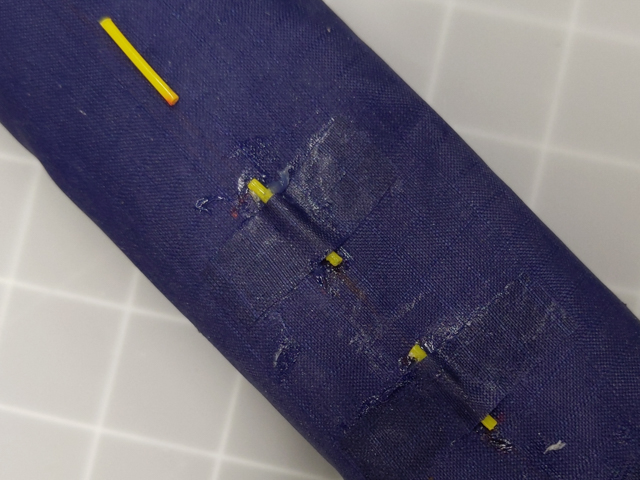}}
  \subfigure[]{\includegraphics[width = 0.25\columnwidth]{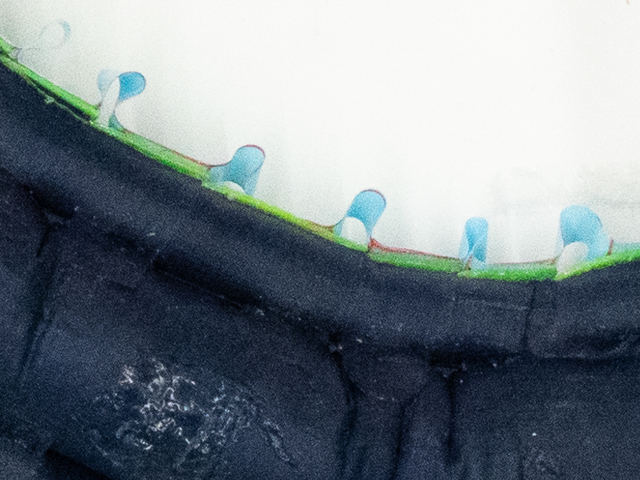}}
  \subfigure{\includegraphics[width = 0.75\columnwidth]{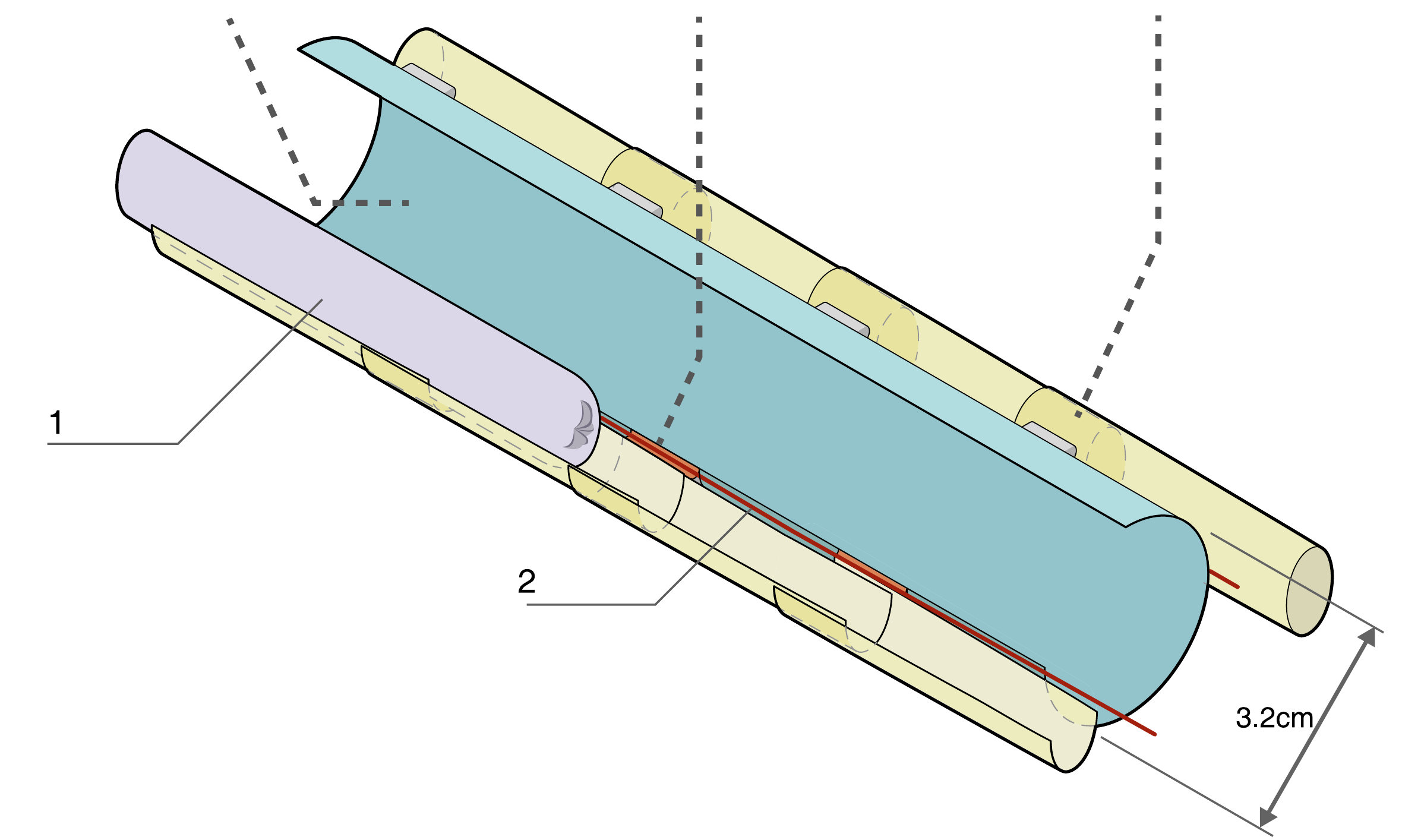}}
  \caption{The composition of the main body of the shape-locking robot is illustrated with a cut-away view. The photos shows the construction of the corresponding parts:  (a) the rolled nylon fabric tube before catheter segments are attached; (b) catheter segments added, with the upper left segment not yet attached; (c) Body with a series of guiding tube pieces. Other notable components include: 1, the shape-locking body; 2, nylon tendon.}
  \label{fig:FbricationDetails}
\end{figure}

\subsection{Mechanical Drivetrain and Control}
The growth of the bodies can be controlled by one of two strategies.
In the first, the unspooling reels control the displacement of tip-extension, since growth only occurs when new material is passed from the reel through the body to the tip (Fig. \ref{fig:lockingMechanism}).
However, experiments indicate that the tension exerted at the tip of the robot by the reel causes an undesired beam-buckling effect at the distal portion of the robot.
This effect only occurs in long, thin robot bodies made from supple fabric, which is why this first control strategy was effective for the heavy-duty vine robot presented in~\cite{coad2019vine}. 
Consequently, this control strategy is used only for the locking bodies, which are supported by the main body, and thus do not exhibit the buckling behavior.

Alternatively, we control the extension of the main body by increasing the resistance to growth and modulating the main body pressure.
A 3D printed PLA O-ring squeezes the tail of the main body near the tip, increasing friction and therefore the pressure required to grow, without interfering with the everted portion of the robot.
Because the body will not grow below a threshold pressure of approximately \SI{34.5}{\kilo\pascal}, growth is controlled by modulating pressure above or below this value. 


All motors in the drivetrain are interfaced with a PC via an Arduino microcontroller. While closed-loop control is possible by adding sensors or a camera to the robot, we implement a simple open-loop control on our prototype which is sufficient for demonstrating the shape-locking mechanism.

\section{MODELLING}
\label{sec:model}
This section establishes the models that describe: a) the kinematics of the distal end of the shape-locking robot, b) the forces required for an inflated tip-extending body to maintain a certain bending angle, and c) the friction force supplied by the shape-locking body at various curvatures and internal pressures. Based on the result of (a), we simulate the workspace of a shape-locking vine robot in Section \ref{sec:workspace}, assuming certain physical dimensions and a given number of shape-locking events performed during the growth of the robot. The models derived for (b) and (c) are significant for characterizing the robot’s ability to maintain its configuration.
\subsection{Distal End Kinematics}
\label{sec:kinematics}
\begin{figure}[tb]
  \centering
  \includegraphics[width = 0.8\columnwidth]{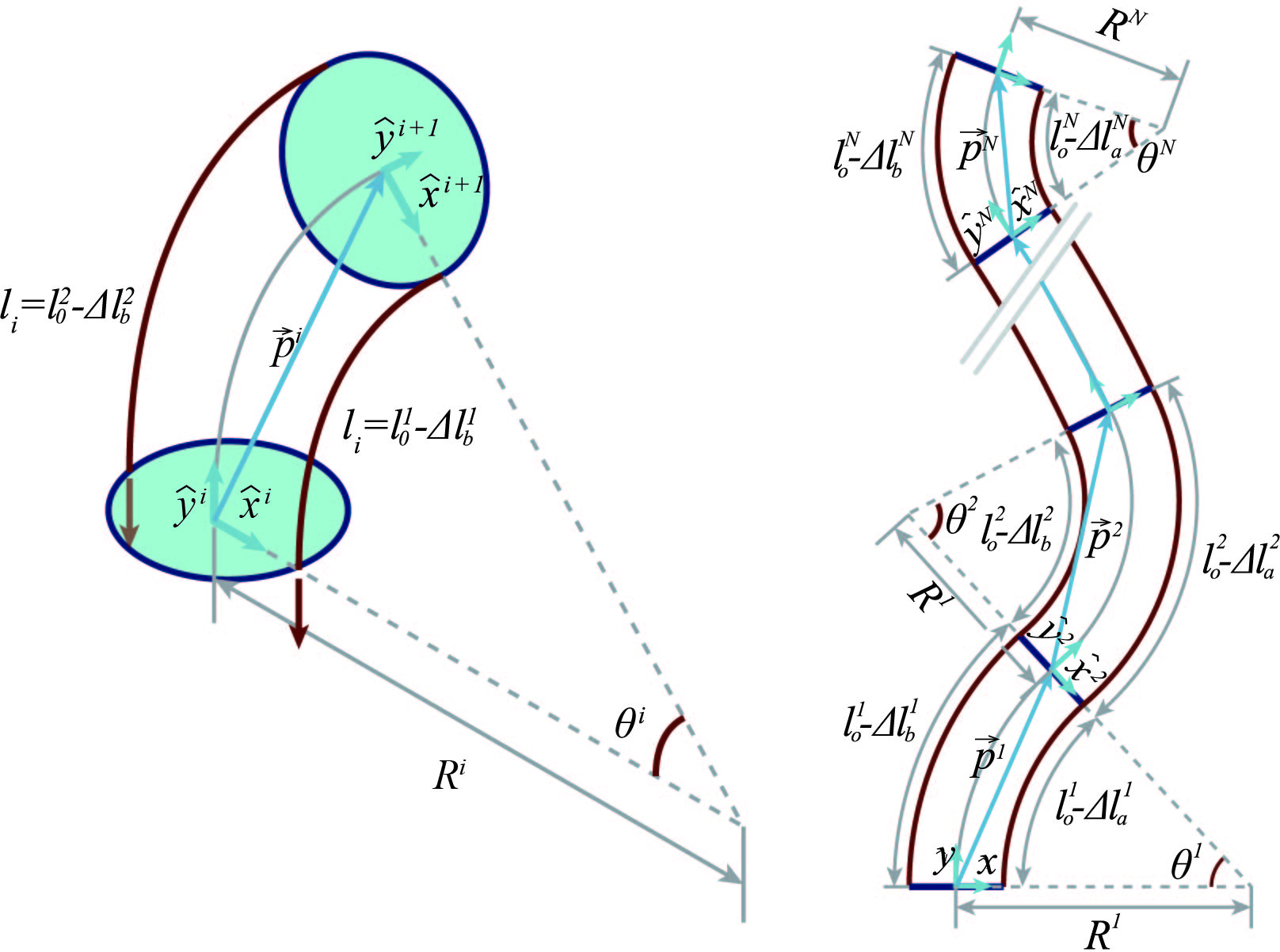}
  \caption{Left: parameters defining the shape of a vine robot segment as described in Section II. Right: kinematic model of a vine robot consisting of N segments resulted from shape-locking.}
  \label{fig:kinematics}
\end{figure}
The shape-locking mechanism can divide the robot into $N$ segments, where $N$ is an arbitrary integer as needed by the desired path of the robot. Each segment has a constant curvature that can be characterized by the length of its internal and external arcs. Given a robot radius $r$, for segment $i$ with a neutral length $l_0^i$ and tendon displacements ${\Delta}l_a^i$ and ${\Delta}l_b^i$ relative to $l_0^i$, we can compute the central angle $\theta^i$ and radius $R^i$ of the curve by
\begin{equation}
    \theta^i = \frac{{\Delta}l_a^i -{\Delta}l_b^i }{2r}
\end{equation}
and
\begin{equation}
    R^i = \frac{2rl_0+r({\Delta}l_a^i +{\Delta}l_b^i)}{ {\Delta}l_a^i -{\Delta}l_b^i }
\end{equation}
as shown in Fig. \ref{fig:kinematics}.

Assuming segment $i$ is parallel to its x-axis, we can compute the position vector $\vec{p}^i$ and rotation matrix $\bm{R}^i$ that describe the rotation of frame $i+1$ relative to frame $i$ as
\begin{equation}
    \vec{p} =
\left [
\begin{array}{c}
    R^i(1-\cos({\theta}^i))\\
	R^i\sin({\theta}^i)
\end{array}
\right  ]
\end{equation}
\begin{equation}
    \bf{R}^i = 
    \left [
    \begin{array}{cc}
    \cos({\theta}^i) & -\sin({\theta}^i)\\
	\sin({\theta}^i) & \cos({\theta}^i)
    \end{array}
    \right  ].
\end{equation}
By making coordinate transformations, we express the end of segment $i+1$ in frame $i$ using
\begin{equation}
    \vec{p}^{i+1}_i = (\bm{R}^i)\vec{p}^{i+1}_{i+1}+{\vec{p}^i_i}
\end{equation}
where the superscript denotes the segment that the endpoint belongs to and the subscript denotes the frame in which the point is expressed. By consecutive transformations, we can obtain the endpoint position of the $N^{th}$ segment relative to the frame attached to the first segment (i.e., the base frame) by
\begin{equation}
    \vec{p}^N_1 = \sum_{i=1}^{N-1}\bm{R}^{i}\ldots \bm{R}^{2}\bm{R}^1\vec{p}^{i+1}_{i+1}+\vec{p}^1_1.
    \label{eqn:coordTrans}
\end{equation}
Furthermore, a quantity of particular interest is the orientation of the robot when its end-effector reaches a particular target. This orientation, expressed as the angle $\phi$ relative to the y-axis of the base frame, can be computed as
\begin{equation}
    \phi = \arccos \left({\bm{R}^{N-1}\ldots \bm{R}^2\bm{R}^1}
\left[
\begin{array}{c}
    sin(\phi^i)\\
	cos(\phi^i)
\end{array}
\right]
\cdot
\left[
\begin{array}{c}
    0\\
	1
\end{array}
\right]
\right).
\label{eqn:phi}
\end{equation}

Equations \eqref{eqn:coordTrans} and \eqref{eqn:phi} are used to find the workspace of the robot.

\subsection{Tendon and Locking Body Force Required for Wrinkling}
\label{sec:tendonForce}

\begin{figure}[tb]
  \centering
  \includegraphics[width=0.8\columnwidth]{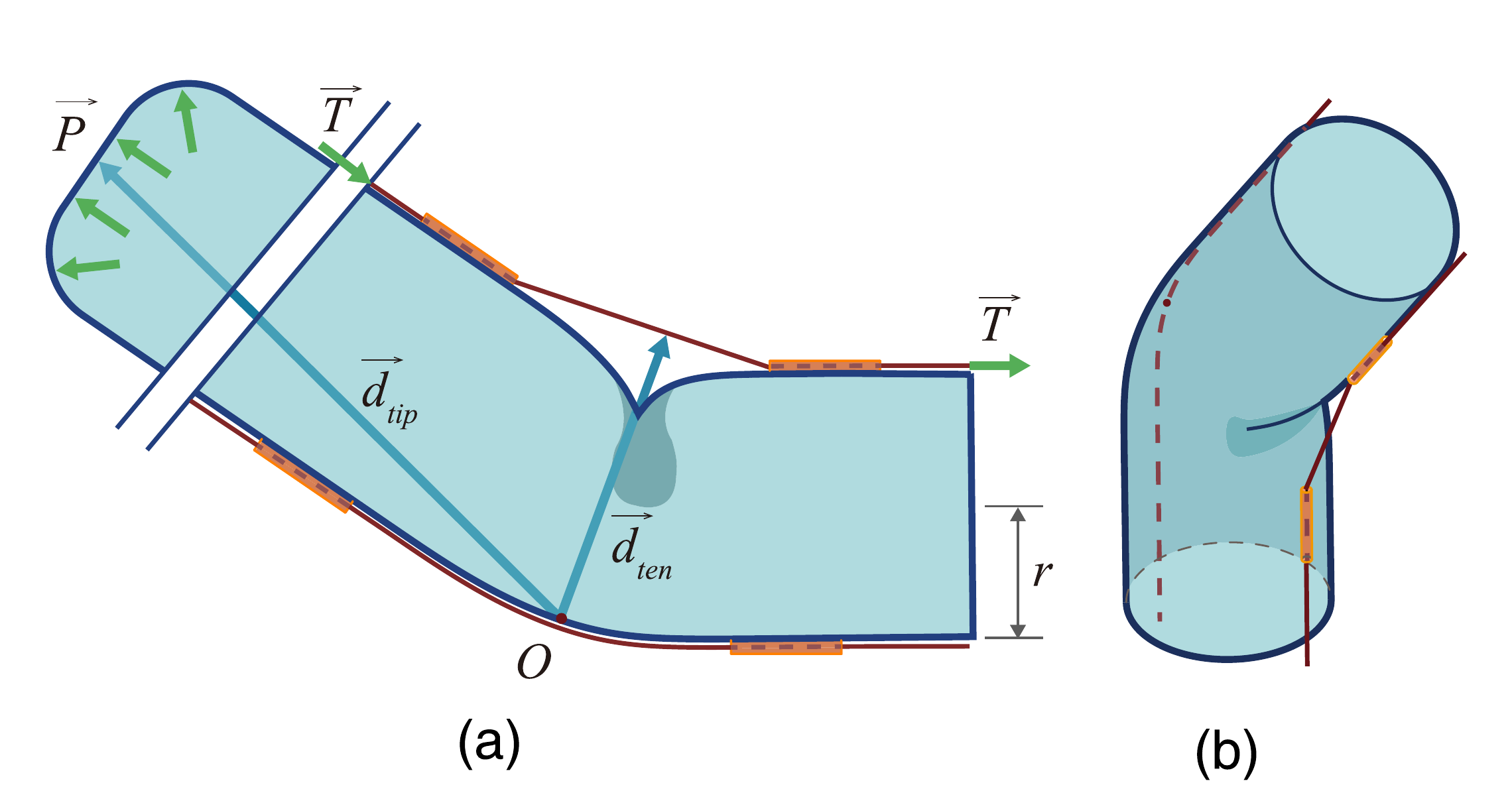}
  \caption{(a): geometry used for wrinkling model. The shaded region indicates the wrinkle. (b): an isometric view of the section described in (a).}
  \label{fig:tendon_moment}
\end{figure}

As illustrated in Fig.\ref{fig:tendon_moment}, the thin-film body of a vine robot wrinkles locally (i.e. a fold forms on the body as a hinge for bending, as described in \cite{levy1963}) to produce a bend. To derive the forces required for maintaining a wrinkle, we begin by taking the moment balance around the midpoint of the arc that is opposite to a wrinkle caused by the contraction of a tendon:
\begin{equation}
    \vec{d}_{tip}\times\pi{r^2}\vec{P}=\vec{d}_{ten}\times\vec{T}.
    \label{eqn:momentBalanceFull}
\end{equation} 
As illustrated in Fig. \ref{fig:tendon_moment}, $\vec{d_{tip}}$ is the position vector from the moment center to the center of the tip cross-section.
$\vec{P}$ is the internal pressure exerted on the tip cross-section, $\vec{d_{ten}}$ is the vector connecting the moment center and the tendon, and $\vec{T}$ is the sum of the tension in the tendon and the locking body.
We assume the locking body lies close enough to the tendon for their lines of action to be considered the same.
A higher tendon tension requires a smaller tension from the locking body to maintain shape. 
For a single wrinkling point that only leads to a small bending angle, we may take the approximation that $|\vec{d}_{tip}\times\pi{r^2}\vec{P}|\approx{\pi}r^3p$ and $|\vec{d}_{ten}\times\vec{T}|\approx2rT$.
While this relationship holds for isotropic body materials, our robot is made from an anisotropic fabric, which is stiff axially and radially, but has a J-shaped stress-strain curve in a direction 45 degrees to these axes.
This leads to fabric deformation at the wrinkling point that is roughly independent of pressure that decrease the wrinkling force. 
This can be represented in our model as a simple offset, $K$.
Accordingly, simplifying \eqref{eqn:momentBalanceFull} and adding this offset yields:
\begin{equation}
    T = \frac{{\pi}{r^2}P}{2}-K.
    \label{eqn:momentBalanceSimple}
\end{equation}

This predicts that the tension required to maintain a wrinkle is approximately constant regardless of angle of deflection if the main body pressure maintains fixed.  

\subsection{Locking Body Friction}
We now consider the friction force, $F_f$, between a locking body and a guiding tube segment. Using an adhesive frictional model due to the soft silicone coating on the locking bodies~\cite{Israelachvili2015Intermolecular}, we have:
\begin{equation}
    F_f = \mu PA + CA,
    \label{eqn:friction}
\end{equation}
where $\mu$ is the coefficient of friction, $P$ is the pressure, $A$ is the area in contact, and $C$ an adhesive constant for the interface. While this model should hold for roughly straight sections, modeling curved sections is more complicated, due to how wrinkling changes the real area of contact and how capstan effects increase friction~\cite{hibbeler_morrow_kokernak_2013}. Therefore, we leave modeling of curved segments for future work.  

\section{RESULTS}
\label{sec:results}
\subsection{Workspace Simulation and Experiments}
\label{sec:workspace}
\subsubsection{Free Space Simulation}
\label{sec:simulationFree}
\begin{figure}[tb]
  \centering
    \subfigure[]{\includegraphics[width = 0.4\columnwidth]{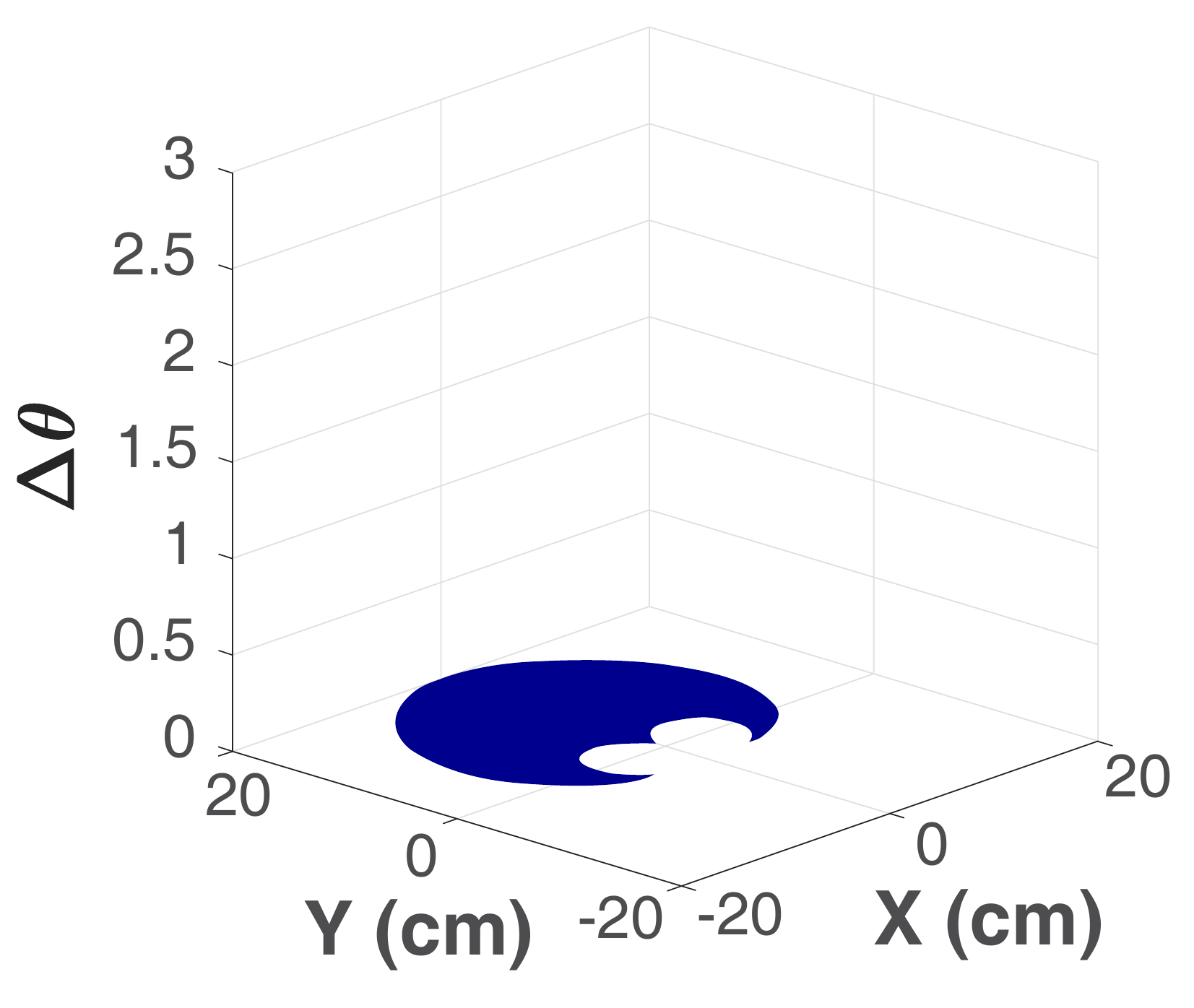}}
  \subfigure[]{\includegraphics[width = 0.5\columnwidth]{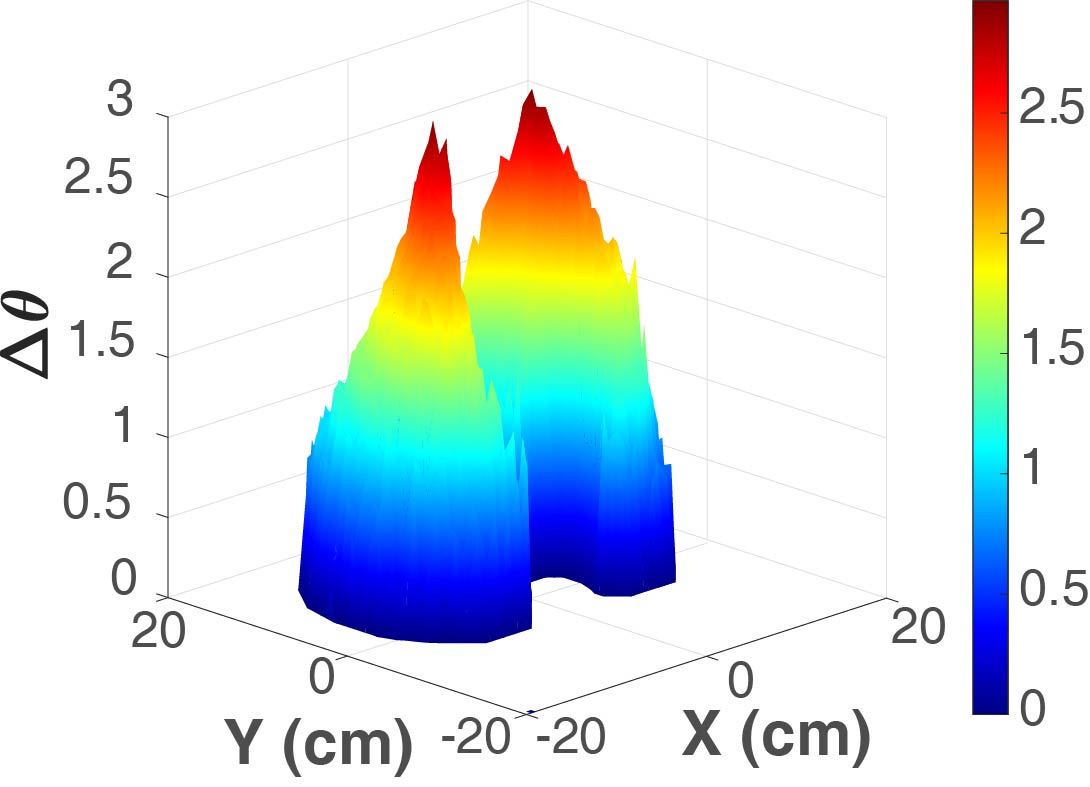}}\\
  \subfigure[]{\includegraphics[width = 0.7\columnwidth]{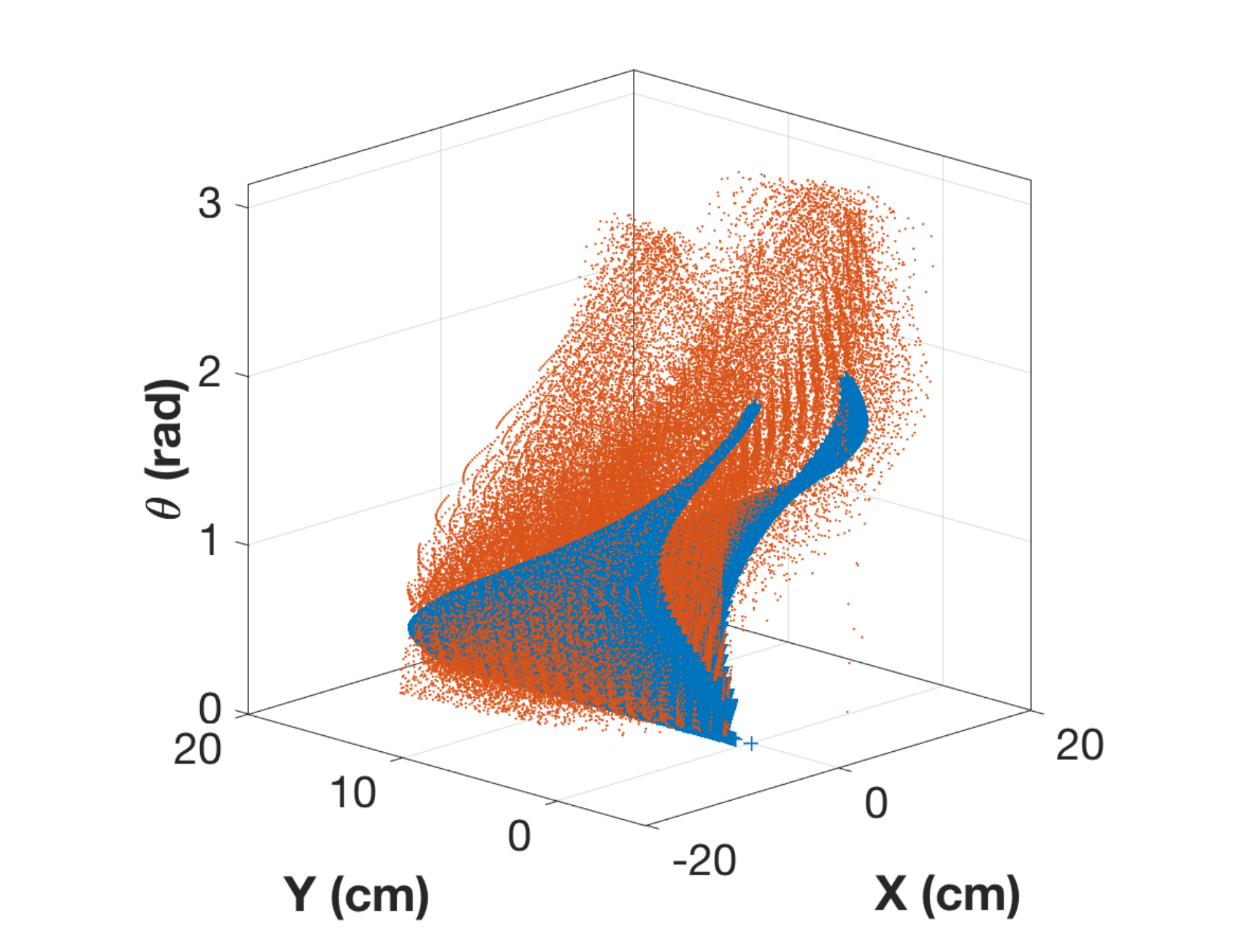}}
  \subfigure[]{\includegraphics[width = 0.2\columnwidth]{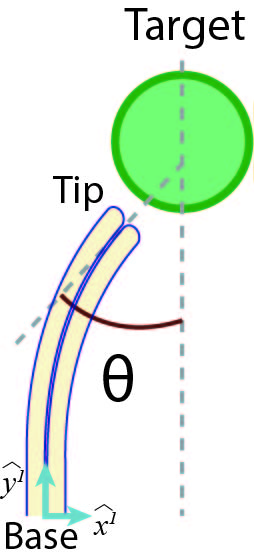}}
  \caption{Workspace of the shape-locking robot in $x$,$y$, and angle of approach $\theta$ in free space. (a) Positions that a vine robot without shape locking can reach. (b) Range of angles that a vine robot with locking can reach with one shape-locking event during growth. The result is a plane because it can only have one orientation at each position. (c) Values (rather than range) of angles that locking (orange) and non-locking (blue) robots can achieve. (d) The definition of $\theta$, $x$ and $y$.}
  \label{fig:workspaceSimulationFree}
\end{figure}

Using the model presented in Section \ref{sec:kinematics}, we numerically simulate the reachable workspace and corresponding range of angles-of-approach for a shape-locking vine robot.
We assume the robot performs shape-locking only once during its extension from start to final length, and that this shape-locking event can occur at an arbitrary length during growth.
The non-locking robot is simply modeled with constant curvature along its length.
Constraints on maximum curvature are applied according to the dimensions of the prototype described in Section \ref{sec:fabrication}. 
Results are shown in Fig. \ref{fig:workspaceSimulationFree}.

\subsubsection{Constrained Space Simulation}
\label{sec:simulationConst}
\begin{figure}[tb]
  \centering
  \subfigure{\includegraphics[width= 0.45\columnwidth]{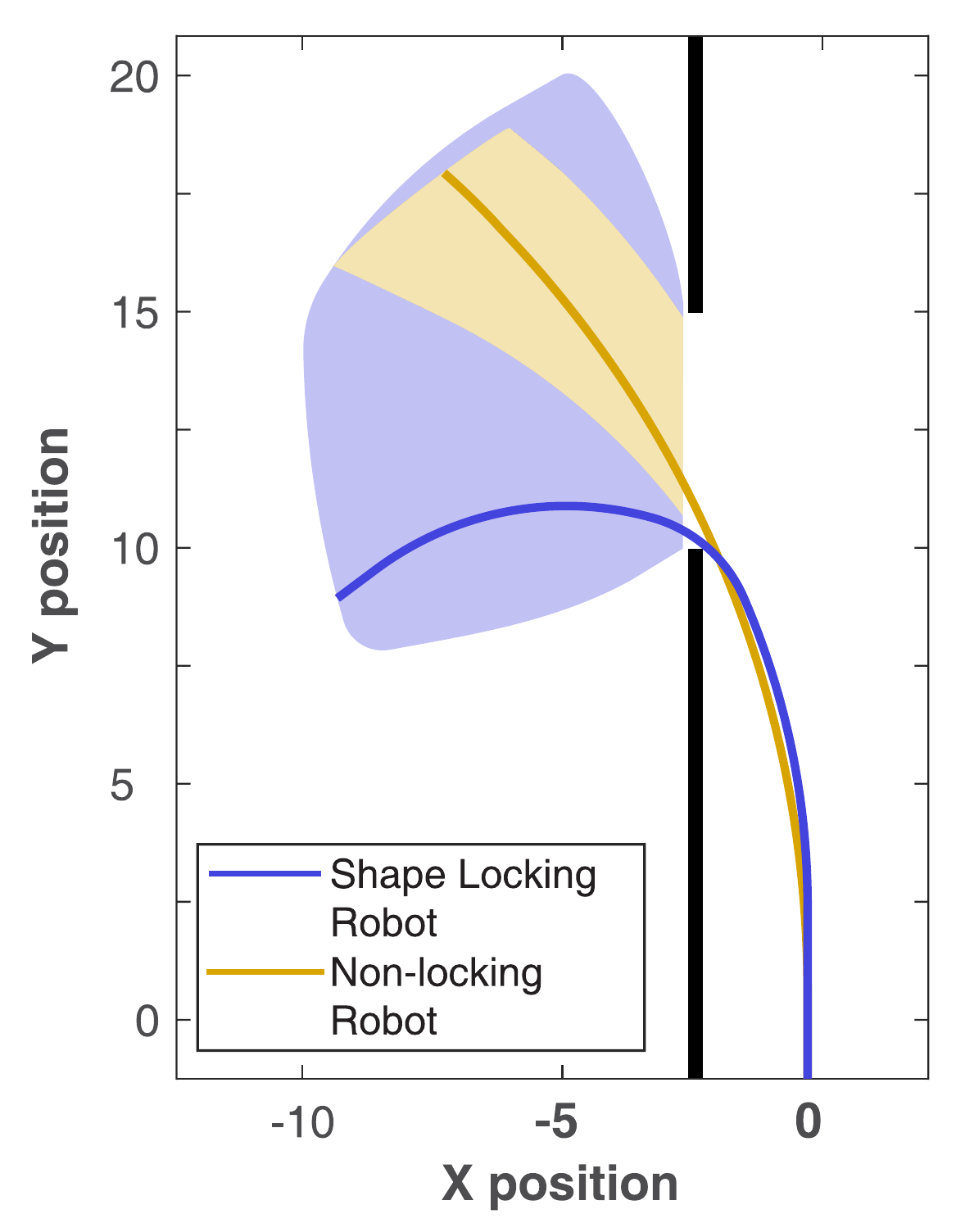}}
  \subfigure{\includegraphics[width = 0.45\columnwidth]{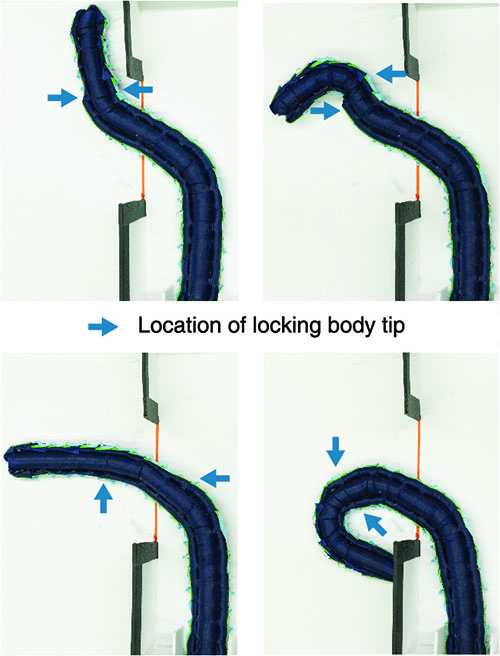}}
  \caption{Left: simulation results showing the workspaces to the left of a vertical wall with an opening for locking and non-locking robots. Two representative body shapes, one for a non-locking robot (orange), and one for a locking robot (blue) are shown. Locking substantially increases the workspace.Left: the shape-locking vine robot navigates through a wall with opening, validating the simulation results.}
  \label{fig:workspaceSimulationWall}
\end{figure}

\begin{figure}[tb]
  \centering
 \subfigure{\includegraphics[width = 0.55\columnwidth]{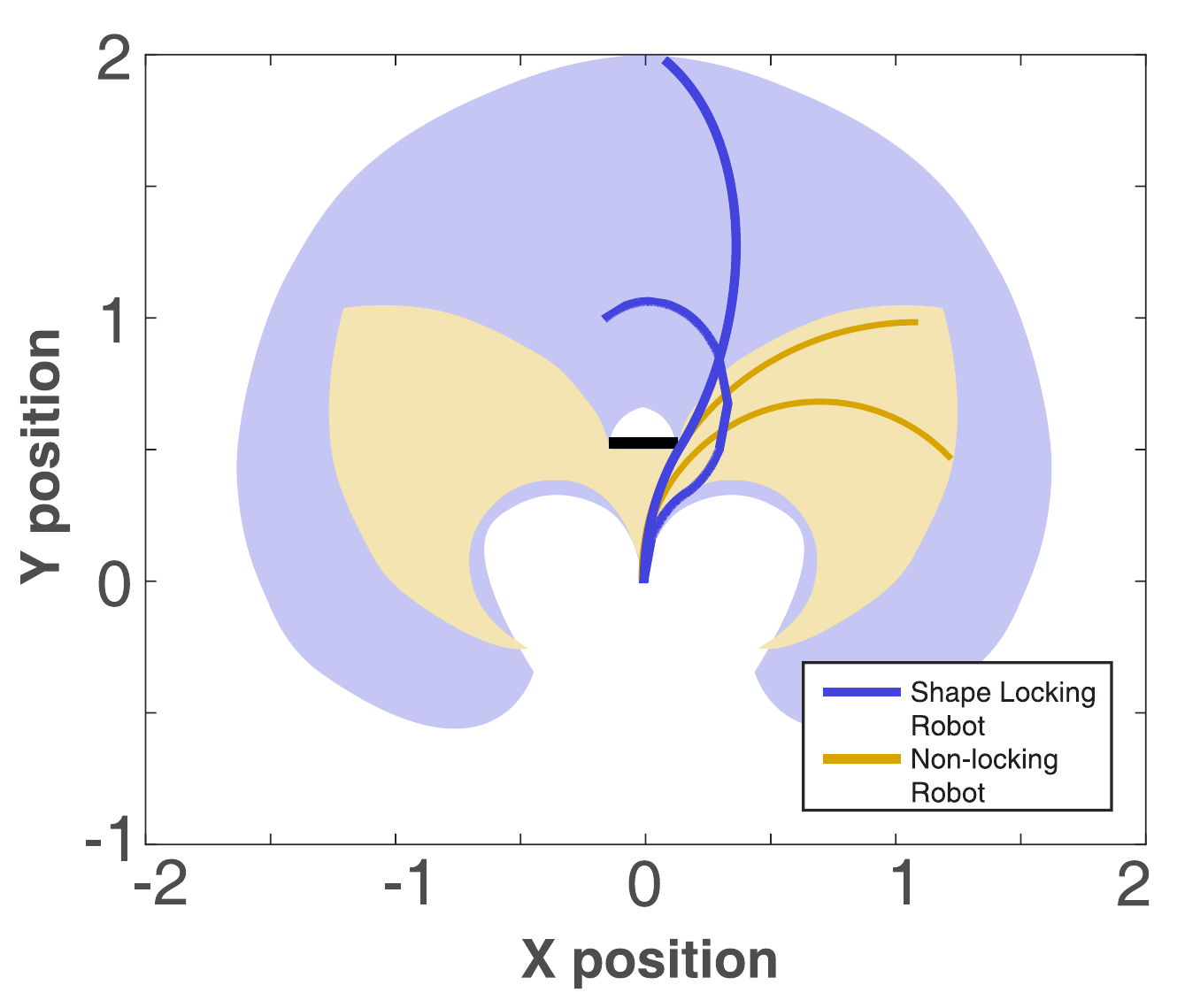}}
  \subfigure{\includegraphics[width = 0.9\columnwidth]{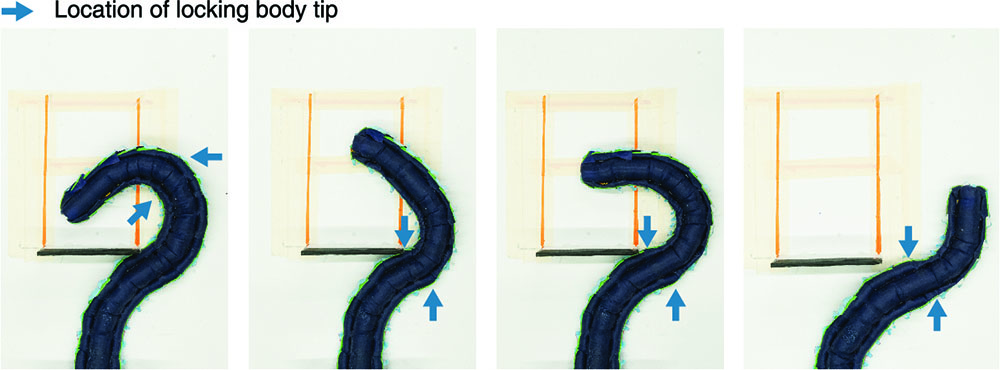}}
  \caption{Up: simulation results showing the workspaces around a horizontal obstacle for both the shape locking robot and a non-locking vine robot. Two representative body shapes, one for a non-locking robot (orange), and one for a locking robot (blue, not grown to the full extent) are also shown. Locking substantially increases the workspace, allowing access behind the obstacle. Note that shape-locking also extends the profile of workspace, as part of the robot can assume a compound curvature that farther the base-to-tip distance. Down: the shape-locking vine robot navigates past a horizontal obstacle, validating the simulation results.}
  \label{fig:workspaceSimulationHorizontal}
\end{figure}

To explore the effect of shape-locking on reachable workspace in constrained environments, we consider two different obstacle arrangements. 
The first obstacle is a gap in a wall parallel to the initial orientation of the robot. 
We model the workspace to the left of the wall, through the opening, in Fig. \ref{fig:workspaceSimulationWall}.
The second arrangement involves a horizontal obstacle directly above the robot's starting location.
We model the the workspace around the obstacle in Fig. \ref{fig:workspaceSimulationHorizontal}.
For both cases, the robot may touch the obstacle but only exert a minimal contact force---that is, the obstacle may not be used to induce passive deformation. 

\subsubsection{Experimental Validation}
To help experimentally validate our model and simulations, we grew the prototype shape-locking robot through environments similar to those in the simulations of Section \ref{sec:simulationConst}. We control the robot to reach the extents of its workspace through a hole in a wall (Fig. \ref{fig:workspaceSimulationWall}), and reach next to and behind a horizontal obstacle (Fig. \ref{fig:workspaceSimulationHorizontal}).   

\subsection{Force Required for Wrinkling}
\label{sec:reslutsTendonForce}
We measured tendon tension vs. robot body deflection  to validate \eqref{eqn:momentBalanceSimple}, the static force model of shape-locking tension.
Because the model only considers a single wrinkling point, only a short section of the prototype was used.
The proximal end of the prototype was mounted to a table, and a force sensor (M3-5, Mark-10 Inc.) mounted on a linear stage was attached to one of the pull tendons.
As the force sensor was pulled away from the prototype, tendon tension increased, causing the prototype to deflect up to \SI{25}{\degree}.
Visual markers on the robot were tracked by an overhead camera at 2 Hz.
The test was repeated with body pressures of \SI{14}{\kilo\pascal} to \SI{41}{\kilo\pascal}.

\begin{figure}[tb]
  \centering
  \includegraphics[width=0.65\columnwidth]{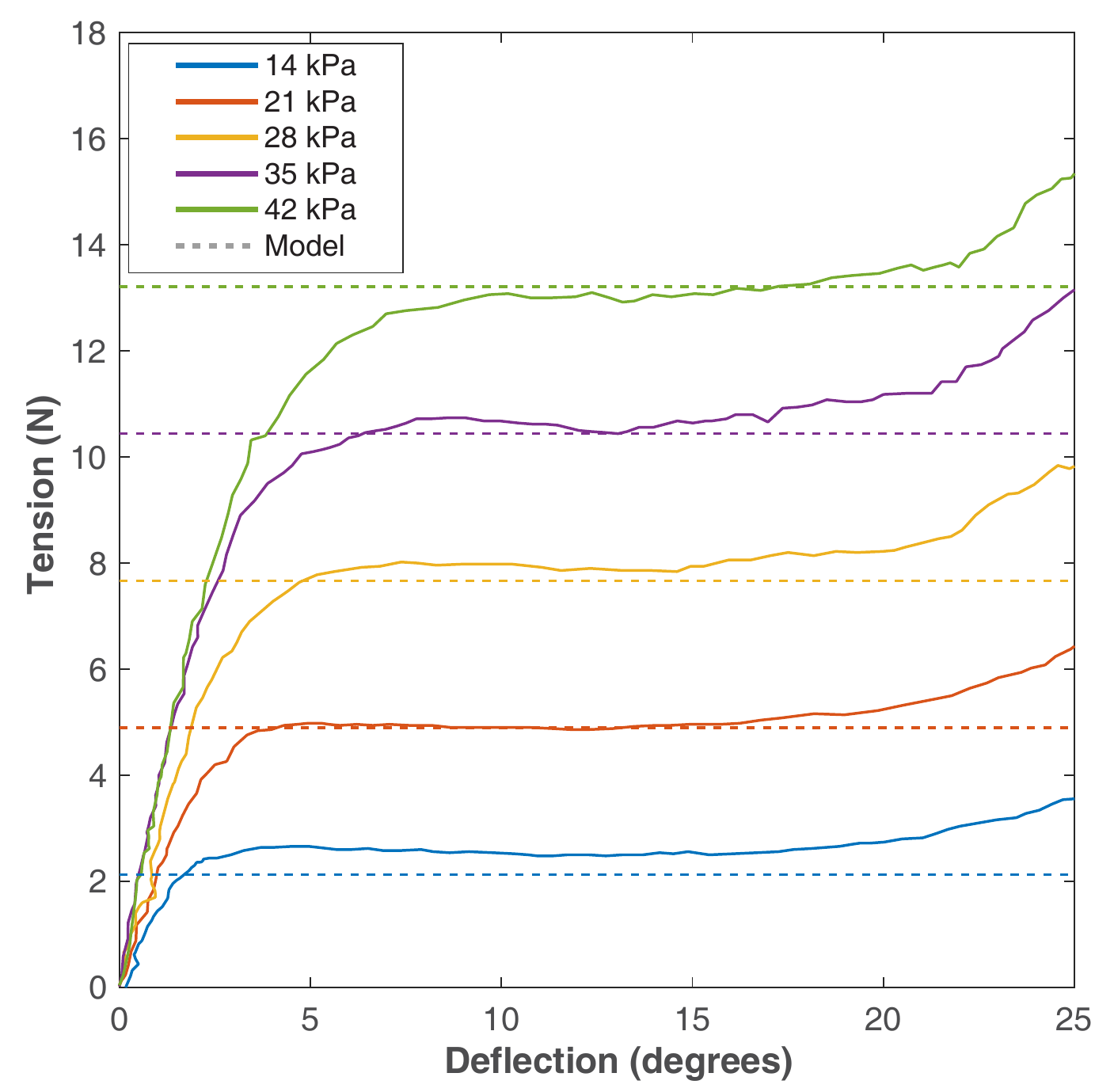}
  \caption{Tension force versus bending angle for a main body tube at varying pressure, with the model from \eqref{eqn:momentBalanceSimple} as dashed lines, with $r$ = 1.6cm and $K$ = 3N. Wrinkling force is nearly constant across a wide range of angles.}
  \label{fig:buckleForce}
\end{figure}

Experimental results are shown in Fig. \ref{fig:buckleForce}, along with the model from  \eqref{eqn:momentBalanceSimple}, with $r$ = \SI{1.6}{\centi\meter} and a least-squares fit offset $K$ of \SI{3}{\newton}.
The data agrees with the model prediction of a near-constant tension across a wide range of deflection angles.

\subsection{Locking Body Friction}
\label{sec:lockingFriction}

The locking body tension required to cause slippage between the locking body and guiding tube was measured to validate \eqref{eqn:friction} for a straight body, and examine the effect of curvature on this frictional force.
Several \SI{100}{\milli\meter} long guiding tube segments were mounted in rigid, curved fixtures with arc angles ranging from \SI{0}{\degree} to \SI{20}{\degree}, and secured at both ends.
A segment of the locking body was then grown into the curved guiding tube with internal pressures ranging from \SI{0}{\kilo\pascal} to \SI{34.5}{\kilo\pascal}, and its proximal end attached to a force gauge (M3-100, Mark-10 Inc.) mounted on a linear stage.
For each curvature, the force gauge was slowly retracted until the locking body first slipped relative to the guiding tube.

\begin{figure}[tb]
  \centering
  \includegraphics[width=0.65\columnwidth]{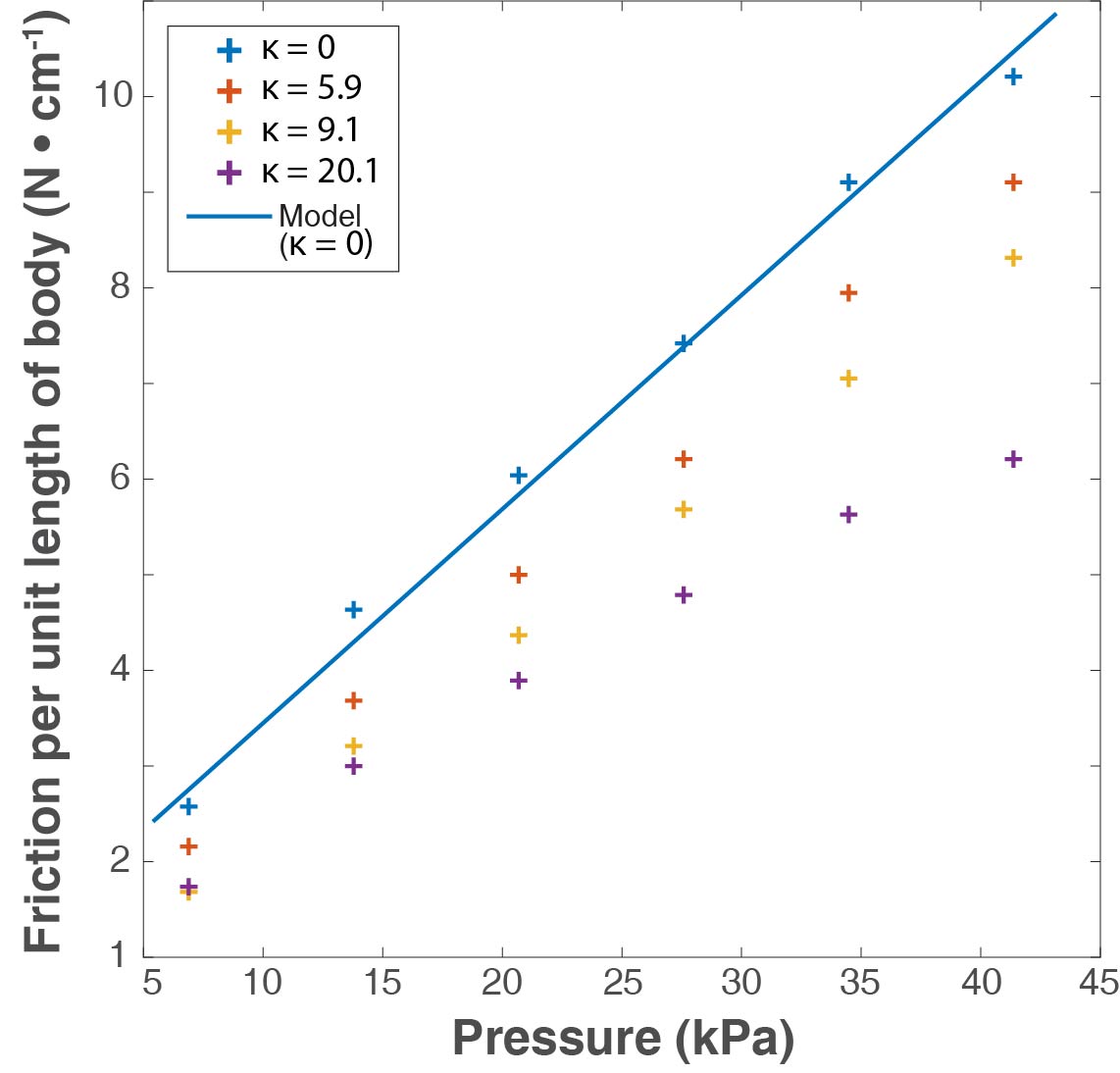}
  \caption{Measured locking body friction (tension per body length) for a range of locking body pressures at various curvatures, $\kappa$, ranging from $\kappa$ = 0 to 20.1 m$^{-1}$. The model from (\ref{eqn:friction}) is shown for $\kappa$ = 0, with $\mu$=0.594 and $C$=0.376\,N$\cdot$cm$^{-2}$.}
  \label{fig:curvature_friction}
\end{figure}

The results are shown in Fig. \ref{fig:curvature_friction}, along with a best fit line for the straight case according to the model in \eqref{eqn:friction}. The values of $\mu$ and $C$ are 0.594 and 0.376N$\cdot$cm$^{-2}$, respectively.
For tests with curvature, the data show a decrease in friction with increasing curvature.
This could be from reduced contact area due to the buckles that necessarily form along the inner surface of the locking body when curved.

\section{DISCUSSION}
\label{sec:discussion}

The advantage of shape-locking is that it allows a vine robot to achieve configurations inaccessible to non-locking vine robots.
Without shape locking, the robot's configuration is limited, and the distal end cannot be manipulated without affecting the rest of the body. 

The simulated workspace in Section \ref{sec:workspace} shows that the locking robot can reach a wide range of orientations at many points in the workspace, whereas a non-locking robot can reach only a single orientation at each point. Further, the simulated and experimental workspaces show that the robot has a larger reachable workspace, in terms of both position and orientation, than a non-locking vine robot in constrained environments. 
This improves the utility of the robot for practical applications. 

The internal force required to deform the robot differs advantageously from a traditional continuum robot.
Our tests in Section \ref{sec:reslutsTendonForce} show that there is a near constant force required to maintain a given curvature.
This contrasts with the increasing force that would be produced by a continuum robot with a linear elastic backbone.
This constant force allows a given locking body pressure to hold a wide range of curvatures.

Section \ref{sec:lockingFriction} shows that the offset linear model of friction holds for straight bodies, giving the friction force that a locking body can produce. 
In conjunction with the tension results in Section \ref{sec:tendonForce}, this can be used to predict the locking body pressure required to maintain a buckle for a given body pressure.
For example, to keep a buckle in a main body at an angle of \SI{15}{\degree} with a main body pressure of \SI{28}{\kilo\pascal} would require a frictional force of about \SI{8}{\newton}, or \SI{4}{\newton\per\centi\meter}, according to Fig. \ref{fig:buckleForce}.
If this buckle is spanned by a locking body segment \SI{2}{\centi\meter} in length and with a bend of \SI{15}{\degree}, its  average curvature $\kappa$ would be 
\SI{13.1}{\per\meter}.
Fig. \ref{fig:curvature_friction} suggests that with this friction requirement and curvature, the locking body pressure needs to be about \SI{20}{\kilo\pascal} or higher.
This pressure would allow the vine robot body to maintain a certain bend while the distal end is manipulated.

The presented prototype and its characterization has a few limitations which will be addressed in future works.
With only two locking bodies and two tendons, it can only be manipulated in two dimensions. 
Adding a third set would allow it to be oriented in three dimensions.
Friction from the pull tendons will also increase exponentially as tortuosity increases due to capstan friction.
This problem could be eliminated by replacing the tendons with soft, pneumatic artificial muscles, such as in \cite{Naclerio2020ral}. 
Furhtermore, while the shape-locking mechanism effectively tunes the stiffness of a vine robot, its modelling and experimental evaluation remains unexplored in this paper. The obstacle configurations considered above, though illustrative, are relatively simple. We thus hope to develop a refined and more general approach to predict the robot's interaction with its environment.




\section{CONCLUSION}
We present a novel concept of shape-locking for a vine-like soft manipulator.
By coordinated growth of the main robot body and two locking bodies, the robot is capable of achieving multiple curvatures along its path, all the while manipulating the distal tip.
This improves its workspace and available orientations over those of a non-locking vine robot, even in the presence of obstacles.
Future, more refined, shape locking vine robots could be used to reach targets in difficult environments, or reduce the interaction between the robot and a sensitive environment.






\balance
\bibliographystyle{IEEEtran}
\bibliography{myReferences}

\end{document}